%% file: HM_2016_Deform.tex
\documentclass[letterpaper,10 pt,conference]{ieeeconf}  
\IEEEoverridecommandlockouts                              
\usepackage{float}
\usepackage{amsmath}
\usepackage{amssymb}
\usepackage{graphicx}
\usepackage{graphics}
\usepackage{blkarray}
\usepackage{booktabs}
\usepackage[pdftex,dvipsnames]{xcolor}  
\usepackage{multirow}
\usepackage{comment}
\usepackage{epstopdf}
\usepackage{tikz}
\usetikzlibrary{arrows,matrix,positioning}
\usepackage[labelformat=simple]{subcaption}
\usepackage{soul}
\usepackage{MnSymbol}
\usepackage{cite}
\usepackage{wrapfig}
\usepackage{units}
\usepackage{xfrac}
\usepackage{algpseudocode}
\usepackage{algorithm}

\graphicspath{{./figures/}}

\usepackage{lipsum}                     
\usepackage{xargs}                      
\usepackage[colorinlistoftodos,prependcaption,textsize=footnotesize]{todonotes}
\newcommandx{\unsure}[2][1=]{\todo[linecolor=yellow,backgroundcolor=yellow!25,bordercolor=yellow,#1]{#2}}
\newcommandx{\change}[2][1=]{\todo[linecolor=Plum,backgroundcolor=Plum!25,bordercolor=Plum,#1]{#2}}
\newcommandx{\info}[2][1=]{\todo[linecolor=OliveGreen,backgroundcolor=OliveGreen!25,bordercolor=OliveGreen,#1]{#2}}
\newcommandx{\improvement}[2][1=]{\todo[linecolor=blue,backgroundcolor=blue!30,bordercolor=blue,#1]{#2}}
\newcommandx{\remove}[2][1=]{\todo[linecolor=red,backgroundcolor=red!30,bordercolor=red,#1]{#2}}

\usepackage{tikz}
\newcommand\copyrighttext{%
  \footnotesize Accepted for presentation in HUMANOIDS 2016, Cancun, Mexico. \copyright 2016 IEEE. Personal use of this material is permitted. Permission from IEEE must be obtained for all other uses, in any current or future media, including reprinting/republishing this material for advertising or promotional purposes, creating new collective works, for resale or redistribution to servers or lists, or reuse of any copyrighted component of this work in other works.}
\newcommand\copyrightnotice{%
\begin{tikzpicture}[remember picture,overlay]
\node[anchor=south,yshift=10pt] at (current page.south) {\fbox{\parbox{\dimexpr\textwidth-\fboxsep-\fboxrule\relax}{\copyrighttext}}};
\end{tikzpicture}%
}

\title{\LARGE \bf
Active Perception and Modeling of Deformable Surfaces \linebreak using Gaussian Processes and Position-based Dynamics
}
\author{ Sergio Caccamo, P{\"u}ren G{\"u}ler, Hedvig Kjellstr{\"o}m, Danica Kragic
\thanks{The authors are with the Computer Vision and Active Perception Lab., Centre for Autonomous Systems, School of Computer Science and Communication, Royal Institute of Technology (KTH), SE-100 44 Stockholm, Sweden.
e-mail: \tt{ $\{$caccamo$|$puren$|$hedvig$|$dani$\}$@kth.se}}}

\begin{document}
\maketitle
\thispagestyle{empty}
\pagestyle{empty}
\begin{abstract}

Exploring and modeling heterogeneous elastic surfaces requires multiple interactions with the environment and a complex selection of physical material parameters. The most common approaches model deformable properties from sets of offline observations using computationally expensive force-based simulators. 
In this work we present an online probabilistic framework for autonomous estimation of a deformability distribution map of heterogeneous elastic surfaces from few physical interactions.
The method takes advantage of Gaussian Processes for constructing a model of the environment geometry surrounding a robot. 
A fast Position-based Dynamics simulator uses focused environmental observations in order to model the elastic behavior of portions of the environment. Gaussian Process Regression maps the local deformability on the whole environment in order to generate a deformability distribution map.
We show experimental results using a PrimeSense camera, a Kinova Jaco2 robotic arm and an Optoforce sensor on different deformable surfaces.

\end{abstract}
\begin{keywords}
Active perception, Deformability modeling, Position-based dynamics, Gaussian process, Tactile exploration.
\end{keywords}

 \copyrightnotice

\input{introduction.tex}

\input{relatedworks.tex}

\input{methodology.tex}

\input{experiments.tex}

\input{conclusions.tex}

\section*{Acknowledgments}
The authors gratefully acknowledge funding under the European Union's seventh framework program (FP7), under grant agreements FP7-ICT-609763 TRADR.

\bibliographystyle{IEEEtran}
\bibliography{InteractivePerception,FieldRobotics,Deformability}

\end{document}

%% file: introduction.tex
\section{Introduction}
\label{sec:intro}

The knowledge of deformability properties of an object or part of an environment can improve robot navigation \cite{frankdeformationmodel} or object manipulation \cite{planningdeformable}. A robot can, for example, avoid unstable terrains while driving, place non-rigid objects on stable positions after manipulation or apply proper forces during grasping.

Visual sensors alone are not enough to extract the level of deformability. Active perception through integration with haptic exploration helps in estimating deformable properties by purposely interacting with and observing the environment. 

Most of the existing methods focus on estimating the deformability of single objects using computationally expensive force based simulators\cite{frankdeformationmodel} and assume that the deformability is homogeneous. Some works consider heterogeneous deformability properties, i.e. deformability is different along the object, 
using a large number of interactions in a complex multi-camera setup\cite{bickelmedical}.

\begin{figure}[t]
  \center
     \includegraphics[width=\columnwidth]{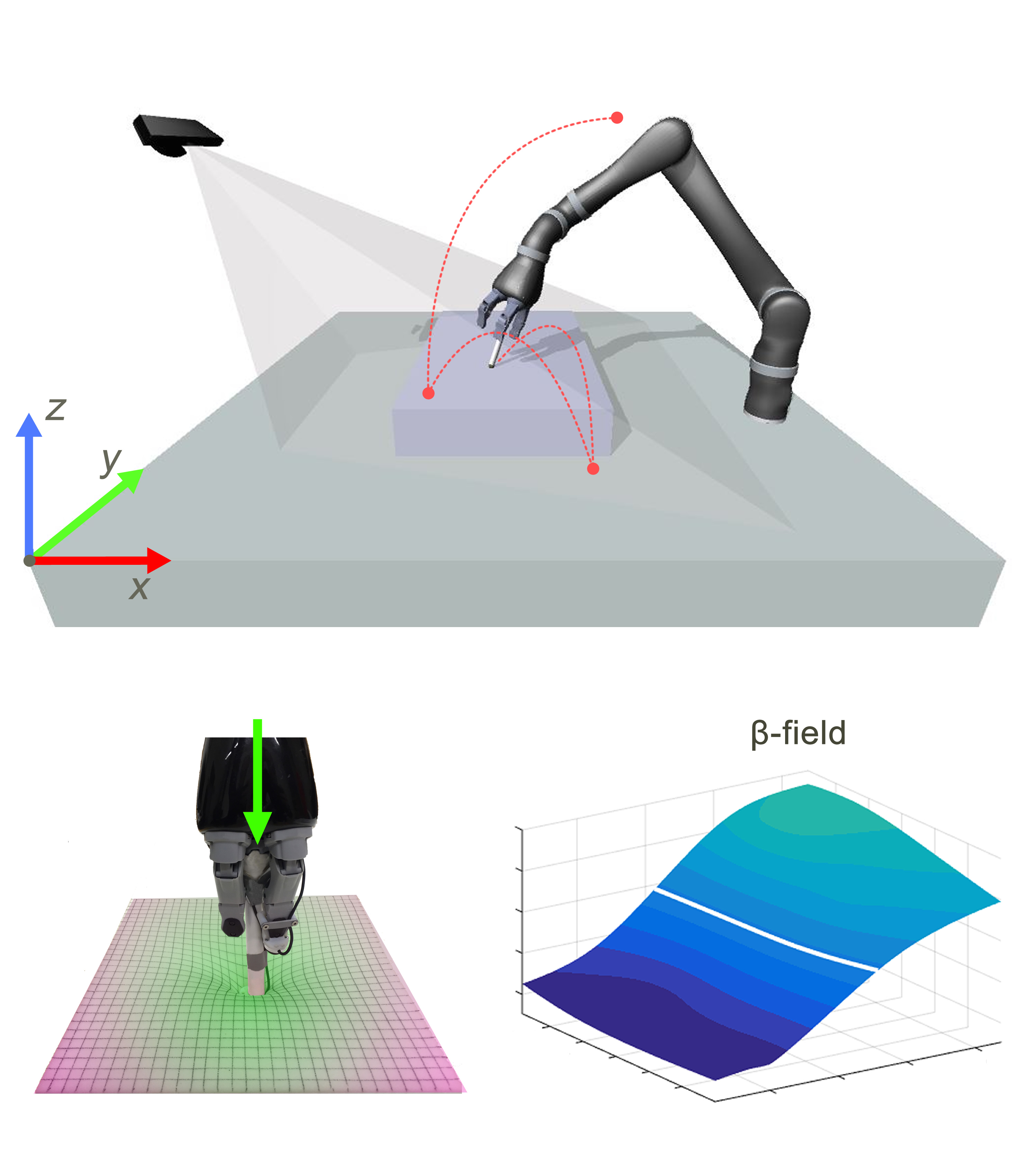}
  \caption{\small{Illustrative representation of the experimental setup. A Kinova Jaco2 arm equipped with Optoforce sensor interacts with a deformable surface observed by a PrimeSense camera. A Gaussian Process models the deformability distribution ($\beta$-field) of the surface from observation and maps it onto the geometric map.}}
  \label{fig:setup_arm}
\end{figure}

We present an active perception framework for extraction of heterogeneous deformability properties of the environment, see Fig. \ref{fig:setup_arm}. The system combines visual and haptic measurements with 
active exploration and builds deformability distribution maps. A fast Position-based dynamics (PBD) simulator is used to estimate the deformability of a portion of surface after a physical interaction.

 We demonstrate the feasibility of our approach through a serie of experiments performed on scenarios representing terrains containing surfaces having different deformabilities.

\subsection{System outline}
\label{sec:systemoutline}

The developed framework follows the process outlined in Fig. \ref{fig:process_flow}. Observations of the environment (initial and final Point Cloud - PC), extracted before and after a physical interactions are used to estimate the local deformability parameters ($\beta$) using a Position-based simulator. The probabilistic model (GPR $\beta$-field) gradually generalizes over the local deformability parameters to build a deformability map of the whole environment. Touch strategy and number of physical interactions are assessed using the joint distributions of the Gaussian Process models.

%% file: relatedworks.tex
\begin{figure*}[t!h]
  \center
     \includegraphics[width=\textwidth]{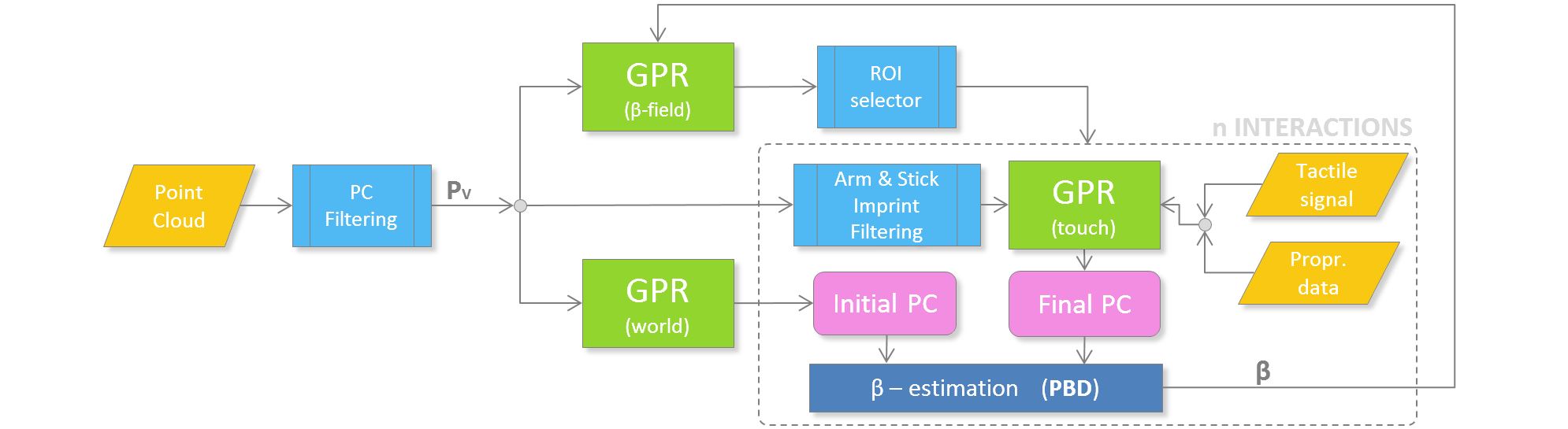}
  \caption{The developed framework proces flow. After a pre-filtering stage, a Gaussian Process Regression (GPR world) is trained and it describes geometry of the environment. A second GPR (GPR $\beta$-field) is used to map deformability parameters ($\beta$ values) onto the world model and determine whether and where to focus the next physical interaction. In each interaction, a new $\beta$ value is locally estimated using a PBD simulator and the GPR $\beta$-field is updated. GPR-touch is used to obtain compact 3D representations of the environment when it is subject to deformation.}
  \label{fig:process_flow}
\end{figure*}

\newpage 

\section{Related Work}
\label{sec:relatedWork}

Gaussian Process Regression (GPR)\cite{rasmussengp} have been widely used for modeling geometric surface properties \cite{callaghangpt} on a broad range of applications such as robotics\cite{dragievgp}, aeronautics or geophysics \cite{Williams1998}.
In robotics, merging visual and haptic sensor data into the same Gaussian Process probabilistic model leads to a better environmental shape representation\cite{yasemingp} or improve planning \cite{dragievgp}.
Environmental observation can condition a GPR so that its posterior mean define the terrain property \cite{marcosgpt} of interest.
Authors in \cite{caccamogpr},\cite{yasemingp} show how to exploit the mean and variance of the joint distribution of a Gaussian Process for enhancing active perception algorithms in modeling geometric properties of objects.

Unlike the previous works, we use Gaussian Process Regression for mapping and modeling the deformability distribution of a surface in an active perception framework.

The problem of modeling the deformation of non-rigid objects have been widely studied in computer graphics \cite{salzman10} and computer vision communities \cite{nealen06}. The most commonly used approaches for modeling deformations are mesh-based models such as finite element method (FEM) and mass-spring model.
FEM aims to approximate the true physic behavior of deformable objects by dividing them into smaller and simpler parts called finite elements. This numerical technique is computationally expensive and has high complexity.
Mass-spring is computationally more efficient than FEM but difficult to tune in order to get the desirable behavior. In recent years, position-based dynamics (PBD) \cite{mullerbsd} have gained attention in the computer graphics community due to their speed and stability. 
PBD based methods converge to the problem solution by solving geometric constraints considering directly the object position and shape. They are computationally efficient, stable and are highly controllable. These are all important assets in the design of a robust and fast active perception framework. Meshless shape matching (MSM) \cite{mullerpbd} is a key algorithm among the field of PBD that simulates rigid and deformable objects\cite{tian13, zhu08, liu08}.
 In this study, we propose to estimate the parameter that define the elastic deformability of the object or part of the environment ($\beta$) from the observed real-world behavior using MSM.  

Estimating parameters of a deformable model is a widely studied approach to realistically simulate the behaviors of objects \cite{bianchi04, bickelmedical}, \cite{frankdeformationmodel, boonvisut13}. Frank et al. \cite{frankdeformationmodel} learn the deformability model of an object by minimizing observed deformations and the FEM model prediction. Also, Boonvisut et al. \cite{boonvisut13} use a non-linear FEM-based method to estimate the mechanical parameters of soft tissues. However, these approaches assume homogeneous material properties. In \cite{bickelmedical}, authors model heterogeneous soft tissues but they rely on a complex experimental setup consisting of several external cameras. 

Unlike the previous approaches, we estimate the deformability of heterogeneous surfaces using MSM and Gaussian Process in a simpler, generic robotic experimental setup, i.e. a robotic arm and a depth camera sensor, see Fig. \ref{fig:setup_arm}. It is showed in \cite{purenpbd} that by matching real-world observation and MSM simulation, the deformability of objects can be estimated in a controlled 2D experimental setup. Here instead, we map the deformability of a surface with heterogeneous material properties by minimizing the error between the model prediction and observed deformations in 3D space.

%% file: methodology.tex
\section{Methodology}
\label{sec:Methods}

In this section, we describe Gaussian Processes for regression (GPR) \cite{rasmussengp} for 2.5 dimensional datasets\footnote{In a 2.5D dataset each $xz$ coordinate has a single height $y$.} (Gaussian Random Field). We discuss how to exploit GPR to generate deformability distribution maps and geometric descriptions and show how to estimate the deformability parameters of an object through observation and simulation. This section ends with a description of the developed algorithms.

\subsection{Gaussian Random Fields}
\label{sec:gaussianprocessesrf}

A Gaussian Process Regression shaped over a bi-dimensional Euclidean set is commonly referred as Gaussian random field.
We start defining the set $\textit{P}_{V} = \{\mathbf{p_1},\mathbf{p_2}\dotsc\mathbf{p_N}\}$, with $\mathbf{p_i} \in \mathbb{R}^3$, of measurements of 3D points generated by the visual sensor system. We define also $D_{RF} = \{\mathbf{x_i},y_i \}_{i=1}^N $ a training set where $\mathbf{x_i}\in \mathbf{X}\subset\mathbb{R}^2$ are the \textit{xy}-coordinates of the points in $\textit{P}_{V}$ and $y_i$ the \textit{z}-coordinates (heights)\footnote{ Axis are described considering the reference frame represented in Fig.\ref{fig:setup_arm}}. Similarly a set $\mathbf{X_*} \equiv \mathbf{X_{{rf}_*}} \subset\mathbb{R}^2$ identifies a set of $M$ test points.
A terrain surface can be described with a function $f:\mathbb{R}^2\to \mathbb{R}$ where each vector of \textit{xy}-coordinates generates a single height.
 This simplistic expression allows to efficiently describe 2.5D terrains but does not allow to model convex shapes which require multiple heights for a single \textit{xy}-coordinate.
  
Such a function can efficiently be modeled by a GPR which places a multivariate Gaussian distribution over the space of $f\left(\mathbf{x}\right)$. The GPR is shaped by a mean function $m\left(\mathbf{x}\right)$ and a covariance function $k\left(\mathbf{x_i},\mathbf{x_j}\right)$. The joint Gaussian distribution, assuming noisy observation $ \mathbf{y} = f\left(\mathbf{x}\right) + \epsilon  \text{ with } \epsilon \sim \mathcal{N}\left( 0, \sigma_n^2\right)$ and  $m\left(\mathbf{x}\right) = 0$ on the test set $\mathbf{X_*}$ assume the following form

\begin{equation}
\begin{bmatrix}
  \mathbf{y} \\
  \mathbf{f_*}
\end{bmatrix}
\sim \mathcal{N}\left( 0,
\begin{bmatrix}
  \mathbf{K} + \sigma_n^2I & \mathbf{k_*}\\
  \mathbf{k_*^T} & \mathbf{k_{**}}
\end{bmatrix}
\right)
\label{eqn:predictivefunct}
\end{equation}
where $\mathbf{K}$ is the covariance matrix between the training points $\left[\mathbf{K}\right]_{i,j = 1 \dotsc N} = k\left(\mathbf{x_i},\mathbf{x_j}\right)$, $\mathbf{k_*}$ the covariance matrix between training and test points $\left[\mathbf{k_*}\right]_{i=1 \dotsc N,j=1 \dotsc M} = k\left(\mathbf{x_i},\mathbf{{x_*}_j}\right)$ and $\mathbf{k_{**}}$ the covariance matrix between the only test points $\left[\mathbf{k_{**}}\right]_{i,j=1 \dotsc M} = k\left(\mathbf{{x_*}_i},\mathbf{{x_*}_j}\right)$.

The predictive function is obtained conditioning the model on the training set \cite{rasmussengp} :
\begin{equation}
p\left(f_* | \mathbf{X_*},\mathbf{X},\mathbf{y}\right) =  \mathcal{N}\left(\overline{f_*},\mathbb{V}\left[f_*\right]\right)
\label{eqn:predictivefunctcond}
\end{equation}

\begin{equation}
\overline{f_*} = \mathbf{k_*^T}\left(\mathbf{K} + \sigma_n^2\mathbf{I}\right)^{-1}\mathbf{y}
\label{eqn:eqmeangp}
\end{equation}
\begin{equation}
V\left[f_*\right] = \mathbf{k_{**}}-\mathbf{k_*^T}\left(\mathbf{K} + \sigma_n^2\mathbf{I}\right)^{-1}\mathbf{k_*}
\label{eqn:eqsdgp}
\end{equation}

We used the popular squared exponential kernel 
\begin{equation}
k\left(\mathbf{x_i},\mathbf{x_j}\right) = \sigma_e^2 \text{exp}\left( - \frac{\left(\mathbf{x_i}-\mathbf{x_j}\right)^T\left(\mathbf{x_i}-\mathbf{x_j}\right)}{\sigma_w^2}\right)
\label{eqn:kernelexp}
\end{equation}
which hyper-parameters $\sigma_e$, $\sigma_w$ were empirically estimated based on a set of experiments made on a 1 m$^3$ area.

The mean of the joint distribution of a Gaussian random field allows to explicitly obtain the heightmap\cite{peckhamdem} of a terrain surface by simply using a grid of bi-dimensional testing points. 
The variance of the random field highlights regions of low density or noisy data, e.g. occluded portion of the map.
In this paper, we use GPR for modeling both the geometric shape of the whole surface under analysis and its deformability properties that we denote $\beta$-field. For the latter, $y_i$ of the training set $D_{RF}$ contains the deformability parameter ($\beta$) of the surface estimated using MSM after a physical interaction on a selected target position.

\subsection{Simulating deformation}
\label{sec:simulatingdeformation}

\begin{figure}[t]
  \center
     \includegraphics[width=\columnwidth]{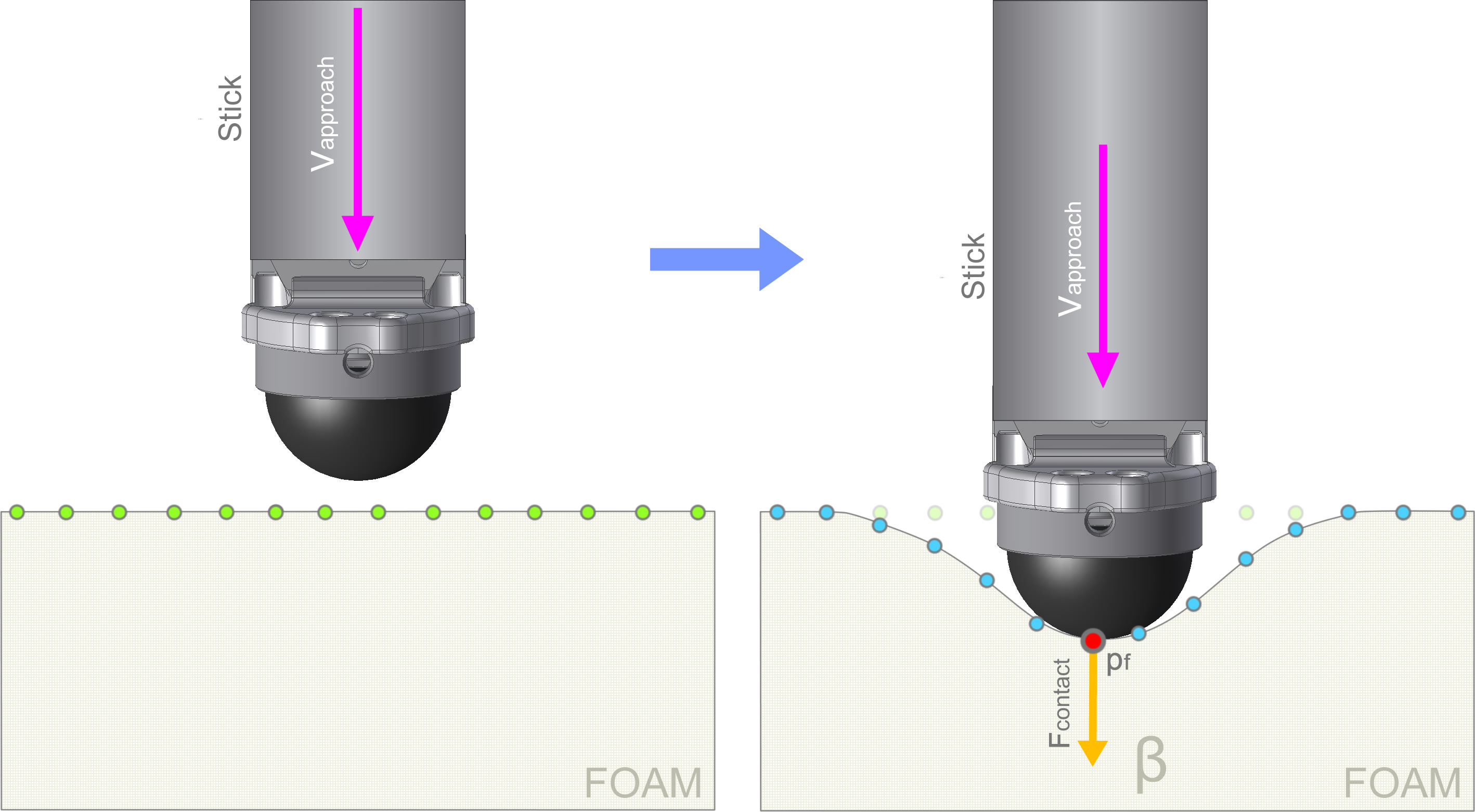}
  \caption{\small{Initial and final reconstructed point cloud used by the Position Based Dynamics algorithm. The Optoforce sensors ensure a constant normal force while collecting the second point cloud. The Gaussian Process Regressions allow to collect grid data points at uniform \textit{xy}-coordinates while filtering noise.  }}
  \label{fig:pointadding}
\end{figure}

\begin{figure}
\centering
\hspace{0pt}
\includegraphics[width=8cm]{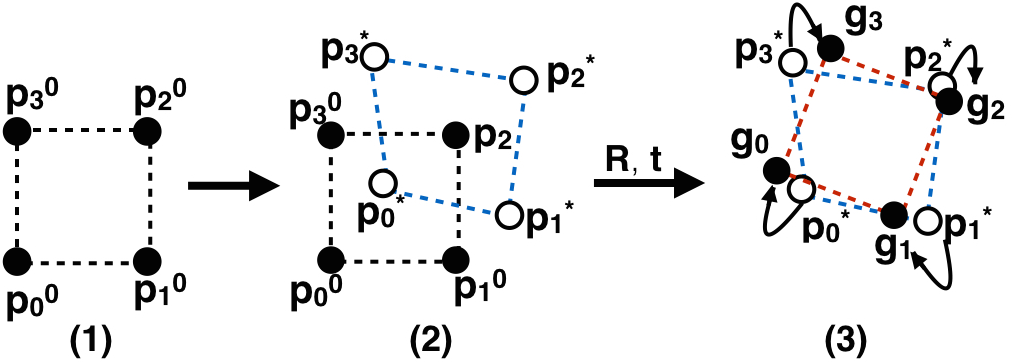}
\caption{For the sake of exposition, let's assume we want to maintain rigidity. (1) The initial shape of the object is represented with the point positions $\mathbf{p}_i^0$. (2) The points are displaced because of external forces and the intermediate deformed shape $\mathbf{p}_i^*$ occurs. The intermediate deformed shape does not embodies the knowledge of the object shape. (3) MSM determines the goal position $\mathbf{g}_i$ by calculating the optimal rotation and translation components that preserves the initial shape. Later, the intermediate deformed points $\mathbf{p}_i^*$ are pulled towards the goal positions $\mathbf{g}_i$. }
\label{fig:basicIdea}
\end{figure} 

The system uses MSM to simulate deformations. The simulation starts by storing the initial shape of the deformable object, $\mathbf{p}_i^0 \in \mathbb{R}^3$ where $i=1, 2, 3, ..., K$ with $K$ the number of points. The basic idea of MSM is shown in Fig. \ref{fig:basicIdea}. 
In each time step, external forces such as gravity or collisions, move the points to unconstrained intermediate deformed positions $\mathbf{p}_i^* \in \mathbb{R}^{3}$.  The unconstrained points are pulled to goal positions $\mathbf{g}_i \in \mathbb{R}^{3}$ which are determined by computing the optimal linear transformation between the initial shape $\mathbf{p}^0  \in \mathbb{R}^{3\times K}$ and intermediate deformed configuration $\mathbf{p}^*  \in \mathbb{R}^{3\times K}$. We then extract the rotational $\mathbf{R}  \in \mathbb{R}^{3x3}$ and translational components $\mathbf{t}  \in \mathbb{R}^3$ of this linear transformation. The rotation and translation are the basis for the rigid transformation that moves the particles towards their goal position which respects the initial shape constraints.

To obtain rotational and translational components, a rotation matrix $\mathbf{R}$ and translation vectors $\mathbf{t}^0$ and $\mathbf{t}$ are determined by minimizing

\begin{equation}
\sum_i m_i (\mathbf{R}(\mathbf{p}_i^0 - \mathbf{t}^0) + \mathbf{t} - \mathbf{p}^*_i)^2~,
\label{eq:min}
\end{equation}
where $m_i$ are the weights of the individual particles. The optimal translation vectors are the centre of mass of the initial shape and the deformed shape:

\begin{eqnarray}
\mathbf{t}^0  =  \frac{1}{m_c} \sum_i^K m_i \mathbf{p}_i^0~, \mathbf{t} = \frac{1}{m_c} \sum_i^K m_i \mathbf{p}^*_i~,  m_c =  \sum_i^K m_i~.
\label{eq:cm}
\end{eqnarray}

Finding the optimal rotation requires more complex steps than finding optimal translation vectors. In \cite{mullerbsd}, authors relax the problem of finding the optimal rotation matrix $\mathbf{R}$ to finding the optimal linear transformation $\mathbf{A} \in \mathbb{R}^{3\times3}$ between the initial configuration $\mathbf{p^0}$ and the intermediate deformed configuration $\mathbf{p^*}$:

\begin{equation} 
\mathbf{A}=\left( \sum_i m_i \mathbf{r}_i {\mathbf{s}_i}^\top \right) \left( \sum_i m_i {\mathbf{s}_i} {\mathbf{s}_i}^\top \right)^{-1} = \mathbf{A}_r \mathbf{A}_s~,
\label{eq:linearTrans}
\end{equation}
where $\mathbf{r}_i = \mathbf{p_i}^*-\mathbf{t}$ and $\mathbf{s}_i = \mathbf{p}_i^0-\mathbf{t}^0$ are the point locations relative to the center of mass. The matrix $\mathbf{A}_s$ is symmetric and contains only scaling information. Hence the rotational part can be obtained by decomposing $\mathbf{A}_r$ into the rotation matrix $\mathbf{R}$ and symmetric matrix $\mathbf{S}$ using polar decomposition $\mathbf{A}_r = \mathbf{R}\mathbf{S}$ as in \cite{mullerbsd}.

We determine the goal position in Fig. \ref{fig:basicIdea} as:
\begin{equation} \label{eq:9}
\mathbf{g}_i = \mathbf{R}\mathbf{s}_i + \mathbf{t}~.
\end{equation}

The steps described in Eq. (\ref{eq:min}-\ref{eq:9}) come from the well known Kabsch algorithm \cite{Kabsch99} and they only allow rigid transformation from the initial shape.
 To simulate deformation, \cite{mullerbsd} introduces linear deformation, e.g. shear and stretching by combining $\mathbf{R}$ and $\mathbf{A}$ as follows:
\begin{equation} \label{eq:8}
\mathbf{g}_i = ((1-\beta)\mathbf{R} + \beta \mathbf{A})\mathbf{s}_i + \mathbf{t}~,
\end{equation}
where $\beta$ controls the degree of deformation, ranging from 0 to 1. If $\beta$ approaches 1, the range of deformation increases, whereas if $\beta$ is close to 0, the object behaves like a rigid body.

$\beta$ is our parameter of interest for defining an object's deformability. Our goal is to estimate it by matching the simulated deformation and the observed deformation of the object. 

Using linear transformation, only shear and stretch can be represented. To extend the range of deformation such as twist and bending modes, quadratic optimal transformation matrix $\mathbf{\bar{A}} \in \mathbb{R}^{3\times 9}$ is calculated as follows and used instead of $\mathbf{A}$ in Eq. \ref{eq:8}:
\begin{equation}
\mathbf{\bar{A}} = (\sum_i m_i \mathbf{r}_i {\mathbf{\bar{s}}_i}^\top)(\sum_i m_i \mathbf{\bar{s}}_i {\mathbf{\bar{s}}_i}^\top) = \mathbf{\bar{A}}_r\mathbf{\bar{A}}_s
\label{eq:quadratic}
\end{equation}
where $\mathbf{\bar{s}}_i = [s_x, s_y, s_z,~ s_x^2, s_y^2, s_z^2,~  s_xs_y, s_ys_z, s_zs_x]^\top \in \mathbb{R}^9$.

\begin{figure}[h]
\centering
\hspace{0pt}
\includegraphics[width=4cm]{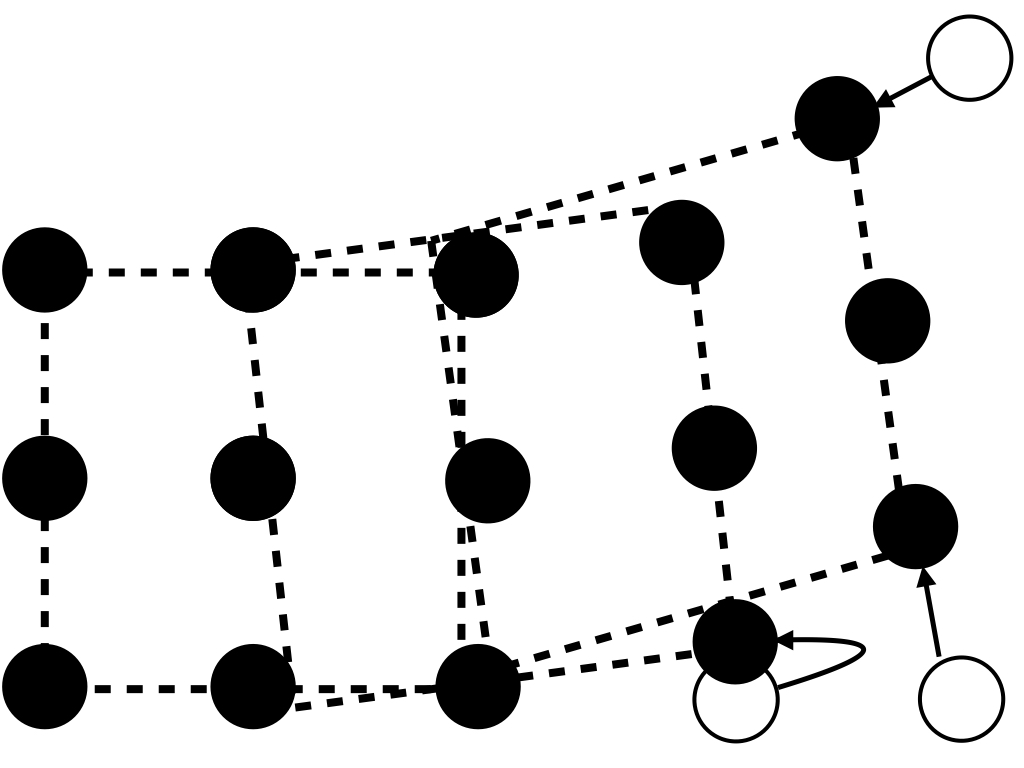}
\caption{ Example point regions configuration with overlapping clusters with size 3x3.}
\label{fig:cluster}
\end{figure} 

For further expanding the range of deformation, the set of points are divided into overlapping clusters as seen in Fig. \ref{fig:cluster} and linear optimal translation $A_j$ of each cluster $j$ is calculated separately.  The size of the cluster was empirically chosen as described in Sec. \ref{sec:results}. At each time step, the final position is determined by blending the goal positions of corresponding clusters:

\begin{equation} \label{eq:cluster}
\mathbf{g}_i = \frac{1}{ M_i } \sum_{j \in \mathfrak{R}_i} \mathbf{g}_i^j~,
\end{equation}
where $M_i$ is the number of clusters that particle $i$ belongs to, $\mathfrak{R}_i$ is the set of clusters particle $i$ belongs to, and $\mathbf{g}_i^j$ is the goal position which is associated with cluster $j \in \mathfrak{R}_i$.

\subsection{Estimating deformability parameter}
\label{sec:betaestimation}

We model the shape of the virtual object as a surface fixed to the ground from edges. The initial shape of the object $\mathbf{p}^0$ is estimated before each physical interaction from the mean of the joint distribution of the GPR as shown in the left side of Fig. \ref{fig:pointadding}.
To simulate the effects of a physical interaction, we select the point $\mathbf{p}_f$ closest to the manipulated region and fix its position in accordance with the disturbance as shown in the right side of Fig. \ref{fig:pointadding}.
The simulator generates a goal configuration $\mathbf{g}^{\beta}$ for a specific $\beta$.

To estimate the deformability parameter $\beta$ that best describes the locally deformed surface, we minimize an error function that measures the distance between the observed deformed shape $\bar{\mathbf{X}}$ and the simulated deformed shape $\mathbf{g}^{\beta}$. $\bar{\mathbf{X}}$ consists of the test set $\mathbf{X}_*$ and heightmap $\bar{f_*}$ modelled by GPR as described in Sec. \ref{sec:gaussianprocessesrf}. The error function is calculated as follows:

\begin{equation}
E(\beta) = \frac{1}{K} \sum_{i=1}^{K} \min_{\mathbf{\bar{x}}_j \in \mathbf{\bar{X}}}({\|\mathbf{g}_i^{\beta} - \mathbf{\bar{x}}_j \|})
\label{eq:minBeta}
\end{equation}
where $\mathbf{\bar{x}}_j$ and $\mathbf{g}_i^{\beta}$ are the $j$th and $i$th points from $\bar{\mathbf{X}}$ and $\mathbf{g}^{\beta}$ respectively. To find the minimum, the simulation runs for a number of $\beta$ values uniformly sampled from the interval $[0,1)$. The $\beta$ that gives the lowest residual in Eq. (\ref{eq:minBeta}) is selected as representing the deformability of the surface. 

\begin{figure*}[t!h]
  \center
     \includegraphics[width=\textwidth]{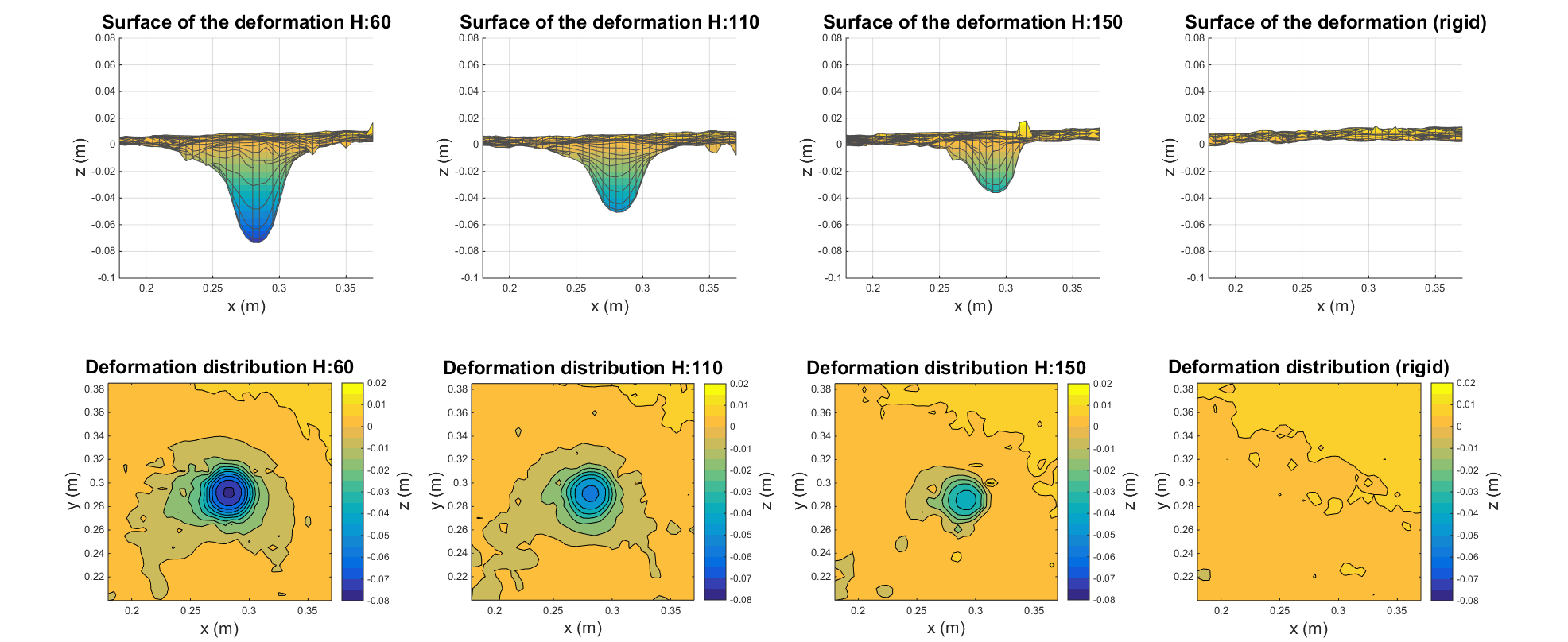}
  \caption{Reconstruction of different deformations of different foams obtained applying the same force on the same contact point. H is the hardness of the surface defined as in Sec. \ref{sec:expscenario}.}
  \label{fig:foam_deformation}
\end{figure*}

\subsection{ The algorithm process flow}
\label{sec:processflowsection}

The active exploration task starts with a full observation of the entire surface under analysis. The point cloud generated from this initial observation is cropped and filtered using a statistical outliers removal filter \cite{statisticoutlierremoval}. A Gaussian random field (GPR world, in Fig. \ref{fig:process_flow}) is trained on the 3D points of the filtered point cloud as described in Sec. \ref{sec:gaussianprocessesrf}. Such GPR builds an internal geometric representation of the environment allowing to obtain compact representations of selected sub-regions (ROI). This is done by considering the mean of the joint distribution of the GPR world model inferred on a dense (0.5 cm) grid of 3D points centered on a ROI.

A second Gaussian Process (GPR $\beta$-field) is then initialized on the \textit{xy}-coordinates of the whole geometric map. 
The block ROI selector of Fig. \ref{fig:process_flow} analyzes the variance of the joint distribution of the GPR $\beta$-field model using a dense (0.5 cm) grid of two dimensional points (\textit{xy}-coordinates) looking for regions of highest uncertainty - meaning that the $\beta$ distribution is poorly modeled because of missing information or high noise. A randomly selected point\footnote{ Selected among the regions carrying higher uncertainty.} is a candidate target for the active exploration task. 

In the successive step, the arm is moved toward the selected region. The approach vector has direction orthogonal to the surface under analysis on the target point location. We use hybrid position-force control \cite{fisherhc} in proximity of the target point to impose a constant force on the direction of the approach vector while allowing displacements along the orthogonal directions. 

When the Optoforce sensor detects a certain normal force the arm stops and the environment is observed again. From this second observation the system generates a new point cloud that contains the local environmental deformation. We use a convex hull filter to remove all the 3D points representing the robot hand and stick. 
The dimension and position of the convex hull is estimated from proprioceptive data using the robot model.
The final point cloud contains occluded regions (incomplete point cloud) because of the hand and stick presence. In order to generate a compact representation of the deformation we train a third GPR (GPR-touch in Fig. \ref{fig:process_flow}) on a squared cropped ROI of the final point cloud that contains the deformation along with 10 tactile points. The tactile points are virtually generated considering the position and shape of the spherical surface of the Optoforce sensor along with the contact force direction as shown in Fig. \ref{fig:pointadding}. From the mean of the joint distribution of GPR-touch we create deformation shapes as shown in Fig. \ref{fig:foam_deformation}. Using the method described in Sec. \ref{sec:betaestimation} we use the two point clouds in order to get a local $\beta$ value. The GPR $\beta$-field is finally trained on the locations subjected to deformation using the estimated $\beta$ value as described in Sec. \ref{sec:gaussianprocessesrf}. The exploration step is repeated until the ROI detector block does not find a new candidate point for the next physical interaction (meaning that the variance distribution of the $\beta$-field is low on the whole map). The threshold value for the variance was empirically estimated through several experiments. Its value determines the numbers of interactions needed and, as we will show in Sec. \ref{sec:results}, the accuracy of the built $\beta$-field map.

%% file: experiments.tex
\begin{figure*}[t]
  \center
     \includegraphics[width=\textwidth]{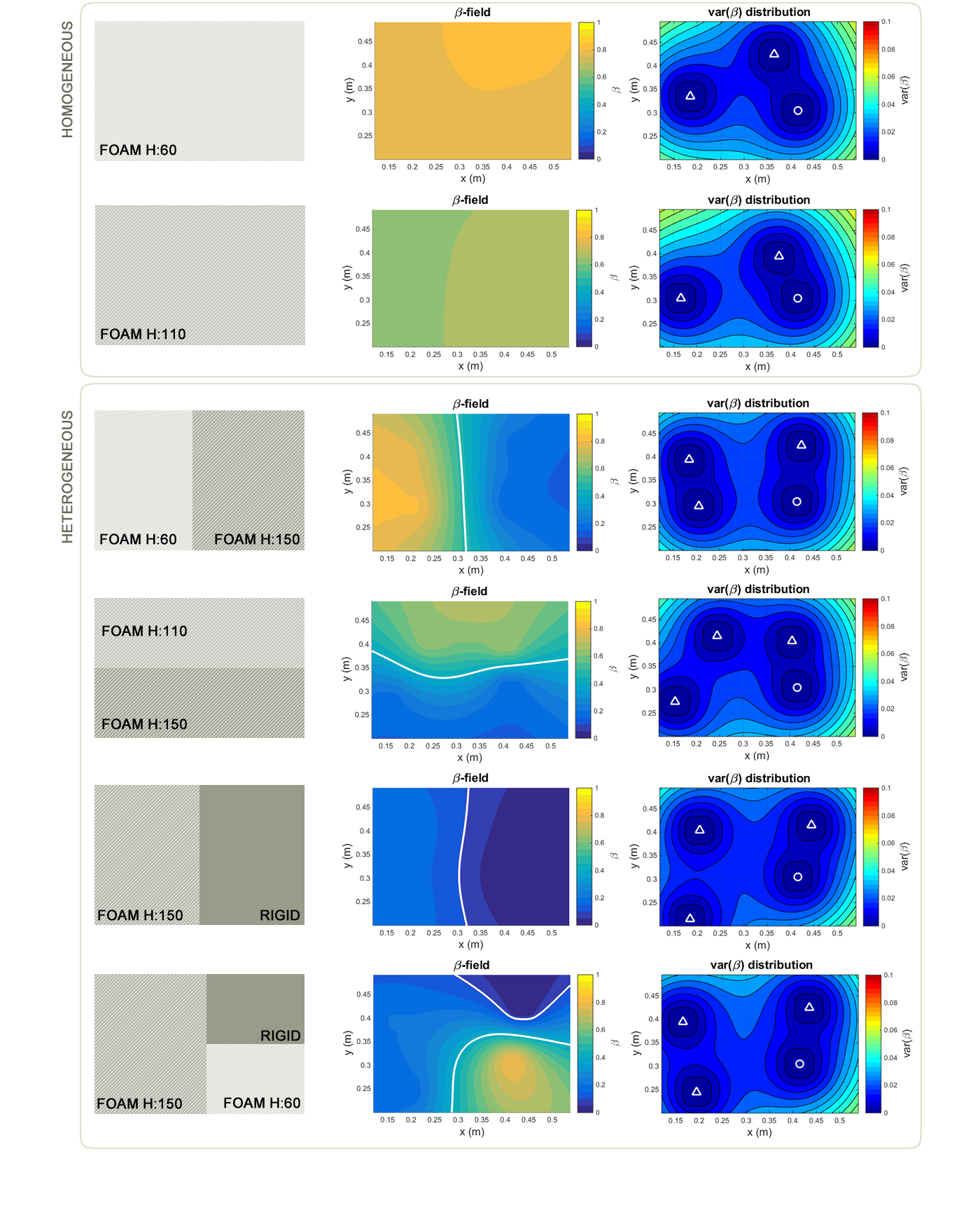}
  \caption{Experiments setup. The first column shows an illustrative representation of the foams size, hardness and position for each experiment (rows). The second column shows the estimated deformability distribution ($\beta$-field). The third column shows the variance of the $\beta$-field along with the contact positions.  }
  \label{fig:resultsbfield}
\end{figure*}

\section{Experimental evaluation}
\label{sec:Experiments}

\subsection{Hardware setup}
\label{sec:expsetup}

The hardware setup (see Fig. \ref{fig:setup_arm}) used in the experiments consists of a PrimeSense RGB-D camera, a Kinova Jaco2\footnote{Kinova website: http://www.kinovarobotics.com/} robotic arm equipped with a 3 fingered Kinova KG-3 gripper and a 3D OptoForce force sensor\footnote{Optoforce website: http://optoforce.com/}.
The camera is placed 80 cm above the table. The relative orientation between the camera and the table plane is 45$\,^{\circ}$. We use a rigid 10 cm stick, mounted on the Kinova hand, for the interaction.

Sets of homogeneous and heterogeneous elastic foams of different shapes are placed on the table and explored by the arm. 
 The OptoForce sensor, that can detect slipping and shear forces with high frequency, is placed on the tip of the stick.
The haptic sensor output consists of a 9D force-position vector generated at 1 kHz. When the desired force is reached, contact force direction together with stick orientation and sensor position (proprioceptive data) are used to generate tactile 3D points.

All the software components (nodes) run under the robot operative system (ROS). Visual data are analyzed using the Point Cloud Library (PCL).

\subsection{Experimental scenarios}
\label{sec:expscenario}

To validate our approach, we tested the framework on six different scenarios described in Table \ref{tab:foam_scenarios}. All the analyzed surfaces covered an area of 60$\times$40 cm. The filtered point clouds covering the analyzed regions counted ${\sim}$12,000 points in average. The foams had different shapes and hardness but equal density. In the first two scenarios, homogeneous foams were physically explored by the arm. In the last four experiments,  several foams having different sizes and hardness were attached together and used to assess the heterogeneous deformability properties. 

The algorithm starts the active exploration task by interacting with a predefined initial $xy$-coordinate. 
 The successive target points were randomly selected among those sub regions of the $\beta$-field having a variance higher than a given threshold.

\begin{figure}[ht]
  \center
     \input{initialImage.tex}
  \caption{\small{
  (a) The optimal $\beta$ as a function of deformability. The estimated deformability increases with decreasing firmness of the surface.  
  (b) The points that represent the shape of the object are divided into overlapping clusters. The cluster size that gives the least error for optimal $\beta$ estimation is selected.
   }}
  \label{fig:betacluster}
\end{figure}
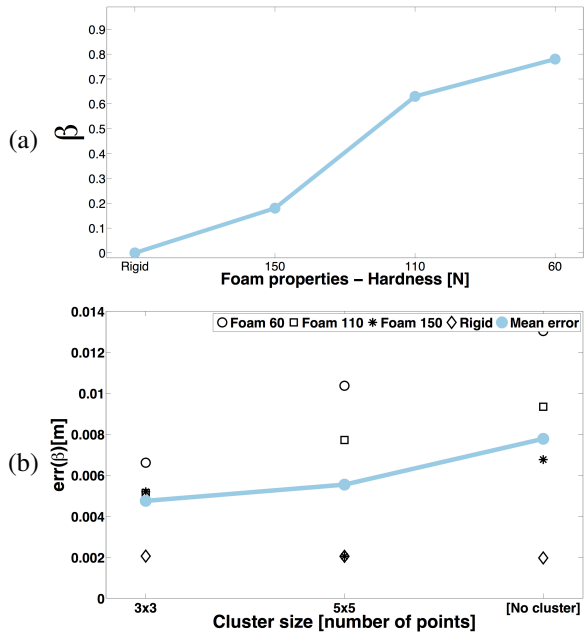

\begin{table}[h]
\begin{tabular}{|l|c|c|c|}
\hline
 Scenario description   & $\text{n}^{\circ}$& $T_h$ var($\beta$) & $\beta$-regions\\
 (hardness)     &  interact.	    &              &     \\
\hline
Homogeneous [60] & 3   & 0.06 & 1 \\
Homogeneous [110] & 3   & 0.06 & 1 \\
\hline
Heterogeneous [60/150] & 4   & 0.06 & 2 \\
Heterogeneous [110/150] & 4   & 0.06 & 2 \\
Heterogeneous [150/rigid] & 4   & 0.06 & 2 \\
Heterogeneous [60/110/rigid] & 4   & 0.06 & 3 \\
\hline
\end{tabular}
  \caption{ Different scenarios setup used during the experiments. The hardness of the foams (H), defined in terms of the force (N) required to compress the foam to 40\%, is provided by the foam manufacturer. $T_h$ represents the threshold value for the variance on the $\beta$-field. }
  \label{tab:foam_scenarios}  
\end{table}  

\subsection{Results}
\label{sec:results}

Fig. \ref{fig:betacluster}(a) shows the deformability values $\beta$, estimated following the approach described in Sec. \ref{sec:betaestimation}, as a function of decreasing hardness. This figure illustrates how the estimated deformability for each foam is in accordance with the real deformation characteristics. In Fig. \ref{fig:betacluster}(b), we also show the evolution of the error as a function of cluster size. The figure indicates that 3$\times $3 cluster size models the observed deformation with the best accuracy. This happens because using small cluster size increases the range of deformation that can be modeled by MSM and therefore the accuracy of the estimation.

Fig. \ref{fig:foam_deformation} shows the mean of the joint distribution of the geometrical random field after contacts with different foams.
It can be seen that the Gaussian random field creates a compact representation of the deformation shape that otherwise would be partially occluded by the stick and affected by noise.

Fig. \ref{fig:resultsbfield} illustrates the interaction steps for the presented scenario and the corresponding evolution of the $\beta$-fields. The first column (ground truth) shows a representation of the setup with the position of the foams along with their hardness and relative dimension. The second column shows the corresponding estimated $\beta$-fields. A contour function identifies the best isopleth with the corresponding subdivisions of the $\beta$-fields into regions. It is possible to notice how the algorithm correctly identifies regions having different deformability and models the whole deformation map consistently with the ground truth. The last column indicates the variance distribution of the $\beta$-field along with the target points selected during the active exploration task. The first target point was pre-assigned and it is indicated with a circle whereas triangles indicate successive contacts.
We invite the reader to note how the subdivision of the $\beta$-field into regions of the last experiment (last row of Fig. \ref{fig:resultsbfield}) slightly differs from its ground truth. This shows a sensibility of the proposed framework to the selected variance threshold. An increase in the variance threshold (which was empirically chosen during our experiments) helps limiting the number of interactions needed (less target points found) but at the same time decreases the accuracy of the $\beta$-field (regions having low variance are considered as explored). All the experiments lasted in average ${\sim}$1.5 min including arm motion, planning and $\beta$-field calculation.

%% file: initialImage.tex
\tikzstyle{background grid}=[draw, black!50,step=.5cm]
\begin{tikzpicture} 
\hspace{0pt}
\node at (0.0,  0.0) (){\includegraphics[trim= 2cm 0cm 0cm 1cm, clip=true, width=8cm]{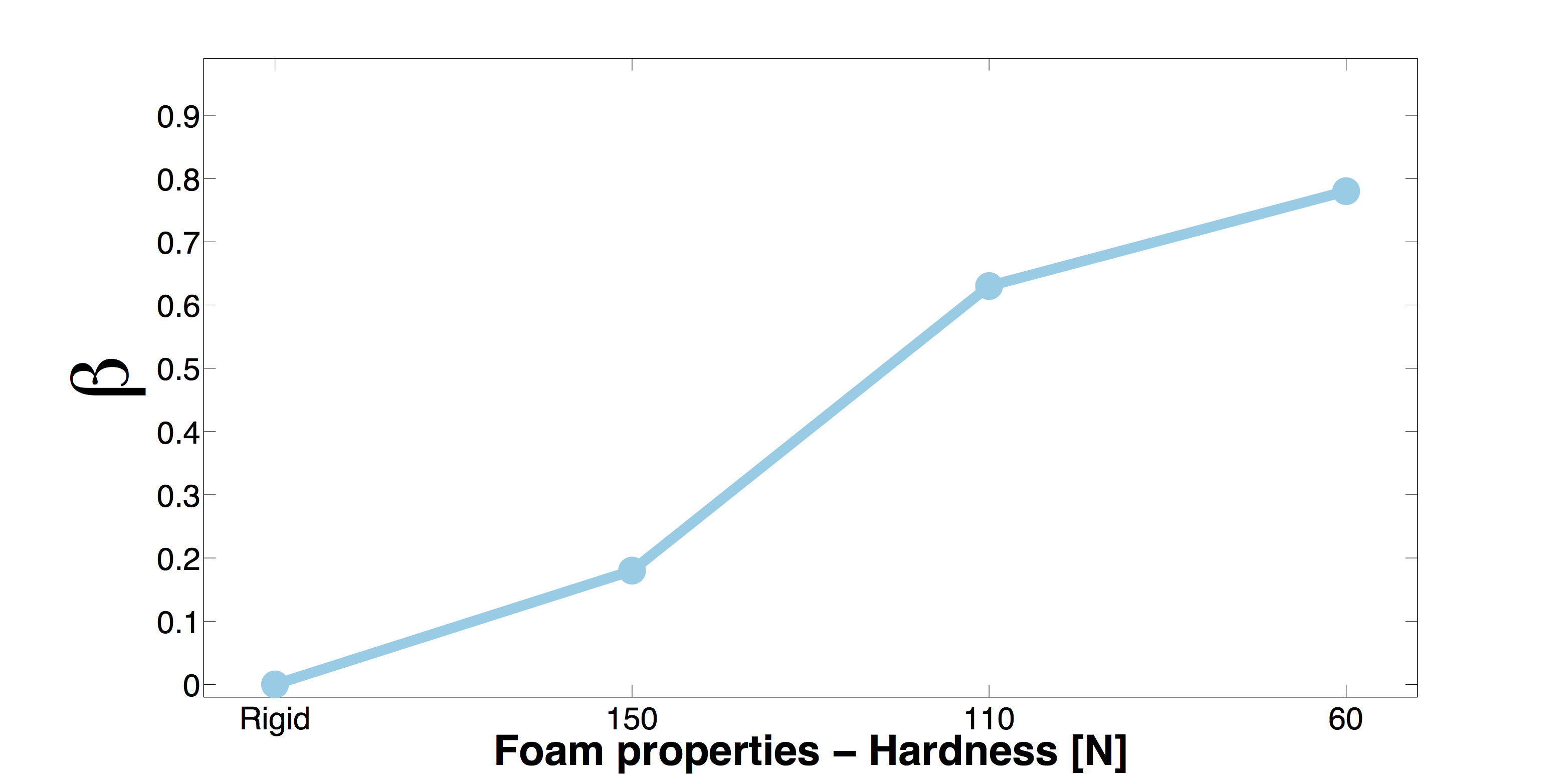}};
\node at (-4.2,  0.0) () {(a)};
\node at (0.0,  -4.3) ()  {\includegraphics[trim= 2cm 0cm 0cm 1cm, clip=true, width=8cm]{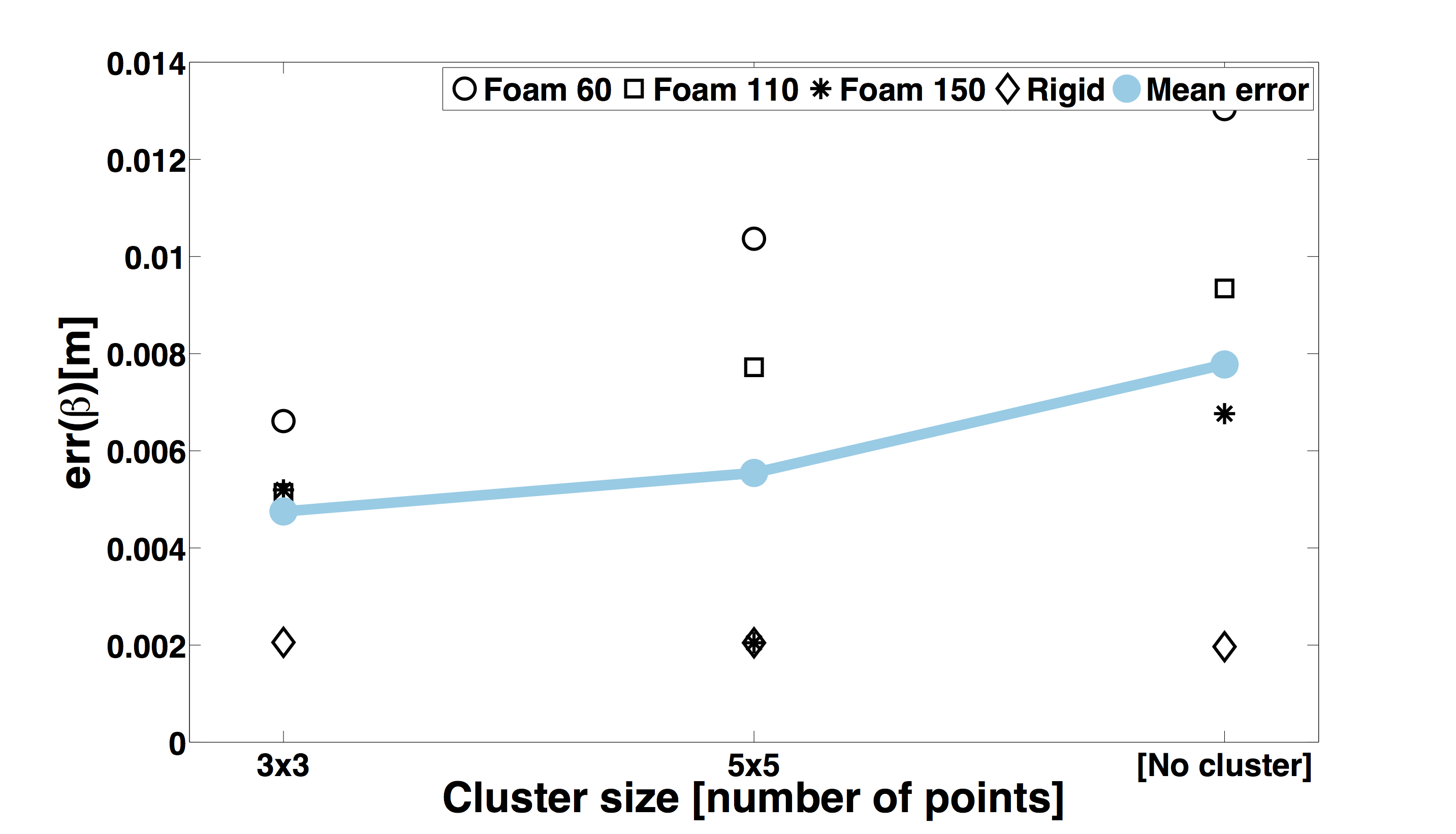}};
\node at (-4.2,  -4.3) () {(b)};
\end{tikzpicture}

 

%% file: conclusions.tex
\section{Conclusions}
\label{sec:conclusions}

We have presented a novel active perception framework for modeling heterogeneous deformable surfaces\footnote{Video of an experiment is available at: https://youtu.be/mDNSDZz7Qzs}. The main contribution of our work is the ability to model the deformability distribution ($\beta$-field) of an environment from few physical interactions.
The novelty of the approach is in the use of real-world observations in a PBD simulator for estimating the deformability parameters.
PBD based methods are computationally efficient which is an important aspect for online active perception tasks.
Our data-driven system relies on multisensory observations and selects regions to be interactively explored for assessing the deformability. 
The presented framework is particularly suitable for applications that require the robot to promptly investigate the environment minimizing the required environmental interactions.
 
We demonstrate the feasibility of our method through several real world experiments, using a simple setup consisting of a robotic arm, an RGB-D camera and a force sensor. We show how the obtained $\beta$-fields of the analyzed surfaces matched the ground truth.

There are several aspects of our method that deserve further attention. We have only modeled elastic, isotropic behaviors of heterogeneous surfaces. We want to increase the potentiality of our framework by capturing plastic and anisotropic behaviors. Another limitation is that the estimated deformability is expressed as a virtual ($\beta$ value) of the deformation rather than a real physical measurement. Such values can change considerably if the simulator settings (e.g. cluster size) change. This can affect the accuracy of modeling surface deformability. Hence the variableness of PBD simulation should be investigated further as a future research. 
Finally, analysis of environmental visual appearance such as color and texture, can help the probabilistic framework to identify regions that are likely to have uniform material properties.